\documentclass{bmvc2k}

\usepackage{multirow}
\usepackage{amsmath}
\usepackage{hyperref}
\usepackage{amssymb}
\usepackage{bbm}

\title{Measuring the Biases and Effectiveness of Content-Style Disentanglement}

\addauthor{Xiao Liu$^{\ *,}$}{Xiao.Liu@ed.ac.uk}{1}
\addauthor{Spyridon Thermos$^{\ *,}$}{SThermos@ed.ac.uk}{1}
\addauthor{Gabriele Valvano$^{\ *,1,}$}{gabriele.valvano@imtlucca.it}{2}
\addauthor{Agisilaos Chartsias}{Agis.Chartsias@ed.ac.uk}{1}
\addauthor{Alison O’Neil$^{\ 1,}$}{Alison.ONeil@mre.medical.canon}{3}
\addauthor{Sotirios A. Tsaftaris$^{\ 1,}$}{S.Tsaftaris@ed.ac.uk}{4}

\addinstitution{
 School of Engineering\\
 University of Edinburgh\\
 Edinburgh, UK
}
\addinstitution{
 IMT School for Advanced Studies Lucca\\
 Lucca, Italy
}
\addinstitution{
 Canon Medical Research Europe Ltd.\\
 Edinburgh, UK
}
\addinstitution{
 The Alan Turing Institute\\
 London, UK\\
 \vspace{3mm}
 $^*$ Equal contribution
}

\runninghead{Liu\protect\etal}{Biases and Effectiveness of C-S Disentanglement}

\def\etal{\emph{et al}\bmvaOneDot}

\begin{document}

\maketitle
\vspace{-4mm}
\begin{abstract}
A recent spate of state-of-the-art semi- and un-supervised solutions disentangle and encode image ``content'' into a spatial tensor and image appearance or ``style'' into a vector, to achieve good performance in spatially equivariant tasks (\textit{e.g.} image-to-image translation). To achieve this, they employ different model design, learning objective, and data biases.
While considerable effort has been made to measure disentanglement in vector representations, and assess its impact on task performance, such analysis for (spatial) content - style disentanglement is lacking. In this paper, we conduct an empirical study to investigate the role of different biases in content-style disentanglement settings and unveil the relationship between the degree of disentanglement and task performance.
In particular, we consider the setting where we: (i) identify key design choices and learning constraints for three popular content-style disentanglement models; (ii) relax or remove such constraints in an ablation fashion; and (iii) use two metrics to measure the degree of disentanglement and assess its effect on each task performance.
Our experiments reveal that there is a ``sweet spot'' between disentanglement, task performance and - surprisingly -- content interpretability, suggesting that blindly forcing for higher disentanglement can hurt model performance and content factors semanticness.
Our findings, as well as the used task-independent metrics, can be used to guide the design and selection of new models for tasks where content-style representations are useful. Code is available at {\small \url{https://github.com/vios-s/CSDisentanglement_Metrics_Library}}.
\end{abstract}

\section{Introduction}
\label{sec:intro}
\begin{figure*}[t]
    \centering
    \includegraphics[width=\textwidth]{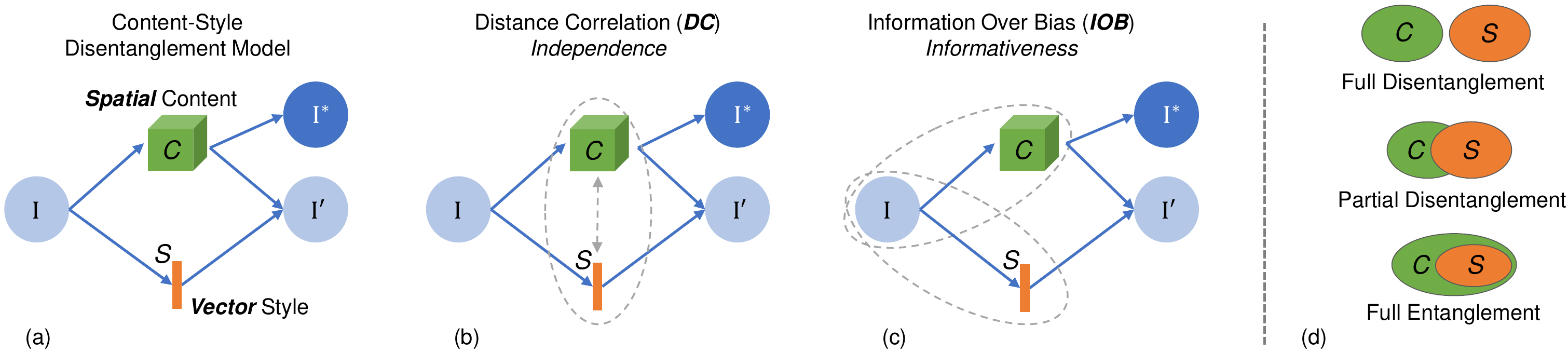}
    \vspace{-6mm}
    \caption{
    \textbf{(a)} A schematic representation of disentanglement between spatial content C and vector style S in the context of a primary and a secondary spatially equivariant task (I$'$, I$^{*}$). 
    Measuring the degree of C-S disentanglement using distance correlation \textbf{(b)} and information encoded over the input bias \textbf{(c)}. \textbf{(d)} A visual description of degrees of C-S (dis)entanglement.}
    \label{fig:main}
    \vspace{-4mm}
\end{figure*}

Recent work in representation learning argues that to achieve explainable and compact representations, one should separate out, or \emph{disentangle}, the underlying explanatory factors into different dimensions of the considered latent space~\cite{bengio2013pami, higgins2018towards}. 
In other words, it is beneficial to obtain representations that can separate latent variables that capture sensitive and useful information for the task at hand from the less informative ones \cite{achille2018emergence}. 
Disentanglement has recently shown to improve task performance, model generalization, and representation interpretability~\cite{desjardins2012arxiv,cohen2014icml,reed2014icml,yang2015nips,kulkarni2015nips,siddharth2017neurips,Esser_2018_CVPR,Esser_2019_ICCV}. 
Unfortunately, disentangling without supervision is an ill-posed and impossible task~\cite{locatello2019iclrw, locatello2020aaai, trauble2020independence} and, to obtain it, we must introduce restrictions and inductive priors~\cite{locatello2019iclrw, locatello2020aaai}. These priors are different forms of ``bias'' imposed by model design (design bias), learning objectives (learning bias), and data (data bias).

In this work, we set out to reveal such choices of bias  in state-of-the-art (SoTA) disentanglement methods. Our particular focus is on ``content-style'' disentanglement, which decomposes input images into spatial ``content'' and vector ``style'' representations. 
In principle, content variables (C) should contain the semantic information required for spatially equivariant tasks (\textit{e.g.}~segmentation and pose estimation), whereas style variables (S) contain information on image appearance (\textit{e.g.}~color intensity and texture). 
However, contrary to extensive research on quantifying the degree of disentanglement between vectors \cite{kumar2018variational,chen2018nips,eastwood2018iclr, ridgeway2018nips,karras2019cvpr,xiao2019arxiv,do2020iclr}, usually there is no analysis of C-S disentanglement. 
In fact, to the best of our knowledge, there is no study identifying the training biases enforced in C-S disentanglement settings or exposing the true relationship between the degree of disentanglement and model performance.
Herein, we attempt to bridge these gaps with our \textbf{contributions}:
\begin{list}{$\bullet$}{}
    \item We identify and analyse the key biases in SoTA models that employ C-S disentanglement. We show how the biases affect disentanglement and task performance (utility) in three popular vision tasks: image translation, segmentation, and pose estimation.
    \item To make a quantitative analysis possible, we propose two complementary metrics building on existing work, to evaluate C-S disentanglement (Fig.~\ref{fig:main}) in terms of amount of information encoded in each latent variable (informativeness) and (un)correlation between the encoded \emph{spatial tensor} content and \emph{vector} style (proxy for independence).
    \item 
    We find that: a) lower C-S disentanglement benefits task performance if a specific style-related prior is not violated; and b) performance is highly correlated with latent variable informativeness. We also assess content semanticness (interpretability).
\end{list}

\section{Related Work}
\label{related}
\textbf{Content-Style Disentanglement.}
Image-to-Image translation has extensively explored the decoupling of image style and content~\cite{unit, drit, munit, kwon2021arxiv}. Content-style disentanglement was also used in other applications, such as semantic segmentation~\cite{chartsias2019factorised} and pose estimation~\cite{charles2013bmvc}, where the content serves as a robust representation for downstream tasks. In general, most methods derive latent spaces capturing C or S information using auto-encoder variants.

These models achieve C-S disentanglement through different biases, such as architectural choices (\textit{e.g.}~AdaIN~\cite{adain2017iccv}, content binarization~\cite{chartsias2019factorised}), learning objectives (\textit{e.g.}~Kullback-Leibler  divergence, latent regression loss, de-correlation losses in vector representations~\cite{song2020arxiv, chang2018scalable}), or supervisory signals (\textit{e.g.}~using content for segmentation~\cite{chartsias2019factorised}). However, the precise effect of each bias on disentanglement and model performance is not thoroughly explored.

\textbf{Evaluating Disentanglement.}
Recently, several methods have been proposed for assessing the degree of disentanglement in a vector latent variable.
A classical approach is \textit{latent traversals}: a visualization showing how traversing single latent dimensions generates variations in the image reconstruction. Latent traversals do not need ground truth information on the factors, and can be used in mixed tensor spaces \cite{chartsias2019factorised, lorenz2019unsupervised} to offer qualitative evaluations. Alternatively, latent traversals can be combined with pre-trained networks to measure the perceptual distance between the produced embeddings~\cite{karras2019cvpr}.

There exist several ways in quantitatively evaluating representations learned by VAEs and GANs. Unfortunately, these methods rely only on vector representations, and some also peruse ground truth knowledge about the latent factors. 
In particular, some methods try to associate known factors of variations (\textit{e.g.}~rotation) with specific latent dimensions~\cite{higgins2017iclr, kim2018icml} or manifold topology~\cite{zhou2021iclr}.
Others measure the ability to isolate one factor in a single vector latent variable~\cite{kumar2018variational}, measuring compactness or modularity~\cite{chen2018nips, eastwood2018iclr, xiao2019arxiv}, linear separability~\cite{karras2019cvpr}, consistency and restrictiveness~\cite{shu2020iclr}, and explicitness of the representation~\cite{ridgeway2018nips}. Lastly, there is work on measuring the factor informativeness in a vector latent variable \textit{w.r.t.}~the input, independence among factors, as well as interpretability~\cite{do2020iclr, eastwood2018iclr}.

The aforementioned metrics cannot be directly employed to C-S disentanglement settings, where the latent factors have different dimensionality (\textit{i.e.}~the style is a vector and the content a spatial multi-channel tensor). However, in this paper we attempt to transfer these concepts to the C-S disentanglement domain, incorporating both spatial (tensor) and vector representations\footnote{Note that the metrics used for our analysis are generic and can be readily applied to vector-based C-S disentanglement methods, such as~\cite{gabbay2020iclr}.} to expand our understanding of the relation between C-S disentanglement and: a) biases adopted by each model; b) task performance; c) representation interpretability.

\section{Measuring Properties of Disentangled Content and Style}
\label{metrics}
Given $N$ image samples $\{I_i\}_{i=1}^N$, we assume two representations of content and style: $\{C_i\}_{i=1}^N$ and $\{\underline{s}_i\}_{i=1}^N$, respectively.
Building on existing work in vector-based disentanglement~\cite{eastwood2018iclr,do2020iclr}, we present two complementary metrics to evaluate two properties in the context of C-S disentanglement: \emph{(un)correlation}, and \emph{informativeness}. We provide evidence that the metrics offer complementary information in Appendix F. Then, we discuss two properties of the disentangled representations, namely their \emph{utility} and \emph{interpretability}.

\textbf{Distance Correlation ($DC$).}
\label{subsec_dc}
Disentangled representations separate content and style into independent latent spaces~\cite{higgins2018towards}, satisfying $p(C,\underline{s})=p(C)p(\underline{s})$. However directly measuring independence between spatial C and vector S with existing metrics is not feasible. Since independent representations must be uncorrelated~\cite{chen2018nips}, we use the \textit{empirical Distance Correlation} ($DC$)~\cite{szekely2007measuring} to measure the correlation between tensors of arbitrary dimensionality. Note that $DC$ is bounded in the $[0,1]$ range, while differently from other correlation-independence metrics, such as the kernel target alignment~\cite{cristianini2002kernel} and the Hilbert-Schmidt independence criterion~\cite{gretton2005measuring}, it has the advantage of not requiring any pre-defined kernels.

For $N$ samples, consider two $N$-row matrices $T_1$ and $T_2$. In general, $T_1$ and $T_2$ row dimension varies as they are formed by concatenating images $I_i$, content features $C_i$ or style features $\underline{s}_i$. For $I_i$ and $C_i$ we first concatenate the channels and then row-scan to form a vector; $\underline{s}_i$ is already a vector.
$DC$ is then defined as:
\begin{equation}
\label{eq:dist_cor}
    DC(T_1,T_2) = \frac{dCov(T_1, T_2)}{\sqrt{dCov(T_1, T_1)dCov(T_2, T_2)}},
\text{ with } \ 
    dCov(X, Y) = \sqrt{\sum_{i=1}^N\sum_{j=1}^N \frac{A_{i, j}B_{i,j}}{N^2}}.
\end{equation}
Here, $dCov$ is the distance covariance between any two $N$-row matrices $X$ and $Y$, while $A$ and $B$ are their respective distance matrices.
In particular, each matrix element $a_{i,j}$ of $A$ is the Euclidean distance between two samples $\|X^i-X^j\|$, after subtracting the mean of row $i$ and column $j$, as well as the matrix mean. $B$ is similarly calculated  for $Y$.
We estimate disentanglement between C and S using distance correlation, $DC(C,\underline{s})$, with values closer to 0 indicating higher disentanglement. 
C and S can be uncorrelated, \textit{e.g.}~$DC(C,\underline{s})=0$, either when they encode unrelated information or when one encodes \emph{all} information and the other encodes \emph{noise}. The latter indicates posterior collapse, thus full entanglement. To tackle this, $DC(C,\underline{s})$ needs a complementary metric to measure the representations' informativeness.

\textbf{Information Over Bias ($IOB$).} 
\label{subsec_info}
To explicitly measure the amount of information encoded in C and S, we introduce the \textit{Information Over Bias} ($IOB$) metric, aiming to detect posterior collapse when C and S are disentangled, but one (C or S) is not informative about the input.
Given latent variables $z \in \{C, \underline{s}\}$ produced from $N$ images at inference, we measure the amount of information encoded in each representation. To do so, we train a decoder $G_{\theta_{n}}$, a neural network with parameters $\theta_{n}$, to reconstruct images $I$, given the features $z$.

Thus, we define $IOB$ as the expectation over the test images of the ratio:

\begin{equation}\label{eq:iob}
    IOB(I, z) = \mathop{\mathbb{E}}_{i}\left[ \frac{\mathrm{MSE}(I_{i}, G_{\theta_{1}}(\mathbbm{1}))}{\mathrm{MSE}(I_{i},G_{\theta_{2}}(z_{i}))}\right] = \frac{1}{N}\sum\nolimits_{i=1}^{N}\left(\frac{\frac{1}{K} \sum\nolimits_{k=1}^{K}||I_{i}^k-G_{\theta_{1}}(\mathbbm{1})||^{2}}{\frac{1}{K}\sum\nolimits_{k=1}^{K}||I_{i}^k- \tilde{I_{i}}^k||^{2}+\varepsilon}
      \right),
\end{equation}

\noindent where $I$ and $\tilde{I}$ are an image and its reconstruction obtained through $G_{\theta_{n}}$; $i=1\dots N$, $k=1\dots K$, $n=1\dots+\infty$ are indices iterating on the test images, the image pixels, and the generator model index (different for each run), respectively; $\varepsilon$ is a small value that prevents division by zero. We justify the above definition of $IOB$ by observing that a post-hoc minimization of the MSE between $\tilde{I}$ and $I$ is equivalent to maximizing the log likelihood (see our analysis in Appendix A).
Note that the ratio aims at ruling out from $IOB$ both data correlations
(common structure, colors, pose, etc., across the images of the dataset) and architectural biases that one could introduce in the design of $G_{\theta_{n}}$. In particular, this is done by computing the ratio between the MSE obtained after training $G_{\theta_{n}}$ to reconstruct the images from their \textit{informative} representation $z$ (\textit{i.e.}~$\mathrm{MSE}(I_{i},G_{\theta_{2}}(z_{i}))$, and after training $G_{\theta_{n}}$ from an \textit{uninformative} constant tensor $\mathbbm{1}$ (\textit{i.e.}~$\mathrm{MSE}(I_{i}, G_{\theta_{1}}(\mathbbm{1}))$). In the latter case, $G_{\theta_{n}}$ will only learn the dataset bias it can model, given $\theta_{n}$. Hence, high values of $IOB$ can be associated with higher information inside the representation $z$, while the lower bound $IOB=1$ means that no information of the images $I$ is encoded in $z$.\footnote{Optimising $G_{\theta}$ with stochastic gradient descent can introduce noise and slightly alter the measure. For example, $IOB$ may, in practice, even be slightly smaller than 1. Thus, we average results across multiple runs and initializations of $G_{\theta}$, which contributes to the computational load of estimating $IOB$.}

\begin{table*}[b]
\caption{Empirical study results for the $DC$ and $IOB$ metrics evaluation using the teapot dataset~\cite{eastwood2018iclr}. Results are in ``mean $\pm$std" format.}
\label{empirical}
\vspace{-2mm}
\footnotesize
\begin{center}
\begin{tabular}{l | r | r | r | r | r}
     \hline
    \multirow{2}{*}{\textbf{Metric}} & GT $C$ & Random $C$ & GT $C$ & Random $C$ & GT $C$ \\
    & GT $\underline{s}$ & GT $\underline{s}$ & Random $\underline{s}$ & Random $\underline{s}$ & Correlated $\underline{s}$\\ 
        \hline
        \multicolumn{1}{l|}{$DC(C,\underline{s})$ ($\downarrow$)}    & 0.17 $\pm 0.00$ & 0.13 $\pm 0.04$ & 0.05 $\pm 0.00$ & 0.13 $\pm 0.04$ & 0.53$\pm 0.02$\\
        \multicolumn{1}{l|}{ $DC(I,C)$  ($\uparrow$)}                & 0.64 $\pm 0.03$ & 0.16 $\pm 0.05$ & 0.64 $\pm 0.03$ & 0.16 $\pm 0.05$ & 0.64$\pm 0.03$\\ 
        \multicolumn{1}{l|}{$DC(I,\underline{s})$  ($\uparrow$)}     & 0.87 $\pm 0.00$ & 0.87 $\pm 0.00$ & 0.04 $\pm 0.00$ & 0.04 $\pm 0.00$ & 0.33$\pm 0.00$\\ 
        \multicolumn{1}{l|}{$IOB(I,C)$  ($\uparrow$)}                & 1.73 $\pm 0.10$ & 1.41 $\pm 0.20$ & 1.73 $\pm 0.10$ & 1.41 $\pm 0.20$ & 1.73$\pm 0.10$ \\ 
        \multicolumn{1}{l|}{$IOB(I,\underline{s})$  ($\uparrow$)}    & 2.47 $\pm 0.78$ & 2.47 $\pm 0.78$ & 0.76 $\pm 0.15$ & 0.76 $\pm 0.15$ & 2.70$\pm 0.26$\\
        \hline
\end{tabular}
\end{center}
\vspace{-3mm}
\end{table*}

\textbf{Utility and Interpretability.}
\label{subsec_util_inter}
As discussed, we can use $DC$ and $IOB$ to measure the degree of disentanglement between latent representations. 
However, one of the primary goals of disentanglement is to improve task performance (utility) and representation interpretability, hence we also investigate the relationship between C-S disentanglement and these two notions.
In particular, we measure utility by quantifying performance on a downstream task, which for disentangled representations is typically image translation~\cite{munit, drit} to translate image content from one domain to another. We also consider tasks using content \textit{e.g.}~to extract segmentations~\cite{chartsias2019factorised} or landmarks~\cite{lorenz2019unsupervised}, and therefore assess how effectively it can be used in downstream tasks. We detail performance metrics for each application in Sec.~\ref{experiments}.

Assessing interpretability is not trivial. Here, we assume that interpretability implies semantic representations. Previously, vector representations were considered semantic if a portion of the latent space corresponded to specific data variations~\cite{infogan, zhu2021and}. Style semantics were qualitatively evaluated with latent traversals of individual dimensions~\cite{chartsias2019factorised}. Thus, we consider a style interpretable if images produced by linear traversals in the style latent space are realistic and smoothly change intensity.
In spatial representations, such data variation should be confined to individual objects: thus, semantic content should split distinct objects into separate channels of C. Wherever possible, we evaluate this with qualitative visuals.

\section{Validating the Effectiveness of $DC$ and $IOB$}
To verify the effectiveness of $DC$ and $IOB$, we design an experiment using the synthetic teapot dataset~\cite{eastwood2018iclr}, which consists of 200k of $64\times64$ pixel resolution images of a teapot with varying pose and colour. Each image of this dataset is generated using 5 ground truth (GT) generating factors (scalars), \textit{i.e.}~azimuth, elevation, red, green, and blue colour, independently sampled from 5 different uniform distributions. We consider the 3 color factors as the GT style (GT S) representation, while as GT spatial content representation (GT C) we leverage the segmentation mask of the object, which correlates with the azimuth and elevation factors (for visual examples see Appendix B).

We first evaluate $DC$ and $IOB$ using the GT C and S representations, and the input images. Then, we sample from a uniform distribution $\mathit{U}[0, 1]$ to generate a new, random style and content representations for each image, and evaluate the metrics using the following scenarios: a) random content, GT style and images; b) GT content, random style and images; c) random content, random style and images. Finally, to approximate the highly entangled C and S scenario, we construct the content-correlated style representations (correlated S) as the azimuth, elevation and red colour factors. For each experiment, we randomly sample 5k images and the GT representations, while all results are the average of 3 different runs. 

\textbf{Results.}~From Table~\ref{empirical}, we observe that for any combination of C and S (except for the correlated S one), the $DC(C,\underline{s})$ is low, which indicates that the representations are highly uncorrelated. This result meets our expectation as the colour (S) and the azimuth or the elevation factors (C) are independent in the teapot dataset. 
However, we also observe a high $DC(C,\underline{s})$ value, \textit{i.e.}~$0.53$, between GT C and correlated S, which verifies that $DC$ can indeed detect the entangled representations case. Additionally, the effectiveness of the $DC$ metric is validated by the high $DC(I,C)$ values when using GT C representations, versus the low values when using random C ones. Note that the $DC$ between the GT S and image is higher than the one between GT C and image, which is reasonable as S and image have nearly one-to-one mapping relationship, while the segmentation masks for different images can be similar. The $IOB$ results, reported in Table~\ref{empirical}, also reflect that the segmentation mask is less informative ($IOB(I, C)=1.73$) about the input image compared to S ($IOB(I, \underline{s})=2.47$) for the GT C and GT S case. This is a result of the strong dataset bias, where given that the object is always a teapot, it is the colour of the reconstructed image that makes it more similar to the input one in terms of MSE. 

\begin{figure*}[t]
    \centering
    \includegraphics[width=\textwidth]{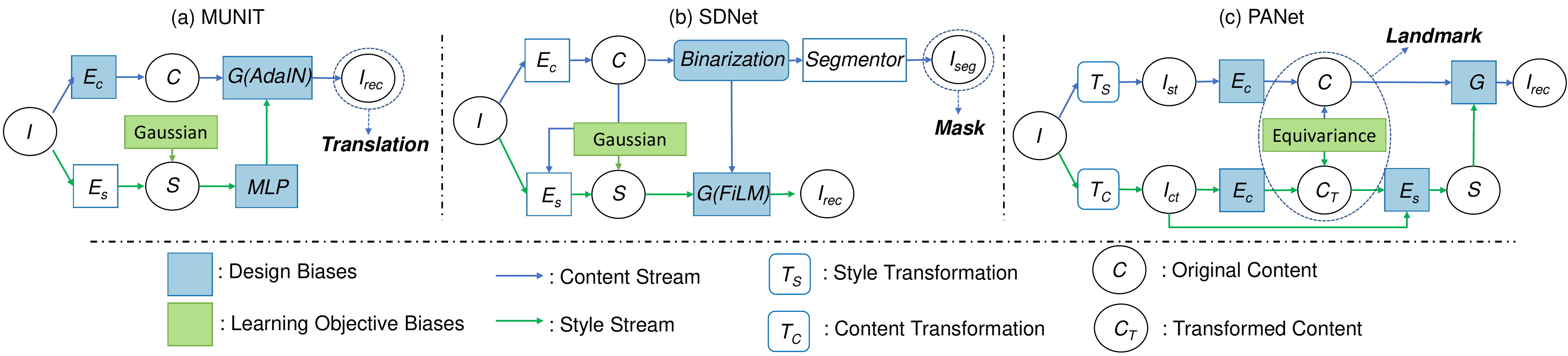}
    \vspace{-4mm}
    \caption{Model schematics. a) MUNIT: Instance normalization is used to remove style from content; $E_{s}$ uses global pooling. b) SDNet: the content is represented with binary features; style is forced to approximate a normal prior. c) PANet: content and style are encouraged to be equivariant to intensity and spatial transformations.}
    \label{fig:models}
    \vspace{-4mm}
\end{figure*}

\section{Experimenting on Vision and Medical Applications}
\label{experiments}
Many applications disentangle C from S~\cite{bouchacourt2018aaai,gabbay2020iclr,park2020neurips,ruta2021arxiv} or other attributes, such as pose, geometry, and motion~\cite{denton2017nips,villegas2017iclr,hsieh2018neurips,xing2019cvpr}, to improve performance in vision tasks.
For our analysis, we select and discuss three SoTA approaches (see Fig.~\ref{fig:models}) from diverse applications, namely image translation (MUNIT~\cite{munit}), semantic segmentation (SDNet~\cite{chartsias2019factorised}), and pose estimation (PANet~\cite{lorenz2019unsupervised}). All resemble auto-encoders, mapping input images to disentangled features but use several biases, which are detailed below. Our scope is to elucidate how each bias affects disentanglement using these models and their chosen biases as exemplars.

Here we describe how each bias is enforced, whilst the detailed model descriptions and a summary of their \textit{design} and \textit{learning} biases can be found in Appendix D.
In particular, for: \textbf{a) MUNIT} we consider ablations removing Instance Normalization (IN)~\cite{ulyanov2016instance}, AdaIN layers, or style Latent Regression (LR) loss (for fairness, we do not remove LR of the content as it is fundamental for the functioning of the model);
\textbf{b) SDNet} we identify content binarization, Gaussian approximation, LR and the FiLM-based~\cite{film2017aaai} decoder as the main biases that affect C-S disentanglement. We investigate their impact on the representations and their effect on semantic segmentation;
\textbf{c) PANet} we remove the Gaussian prior and replace its specific C-S conditioning with AdaIN. We analyse PANet performance in pose estimation. These models help us cover the following diverse cases: \textbf{i)} no supervision and weak C constraints (MUNIT), \textbf{ii)} no supervision with strong C constraints (PANet), and \textbf{iii)} supervision with strong C constraints (SDNet).

\textbf{Setup.}~For each model, we analyze the effect that design choices and learning objectives have on disentanglement and task performance, and we evaluate utility and interpretability of the learned representations. We use the implementations provided by the authors, ablating only the components needed for our analysis\footnote{Metrics code will be made publicly available}. In all tables, arrows ($\uparrow, \downarrow$) indicate direction of metric improvement; best results are in bold. Numbers are the average of 5 different runs. Data description and learning settings can be found in Appendix D (see D.1-D.4).

\begin{table*}[t]
\caption{Comparative evaluation of MUNIT variants using the proposed metrics. We use $FID$ and $LPIPS$ to measure translation quality and diversity between SYNTHIA~\cite{synthia} and Cityscapes~\cite{cityscapes} samples. Results are in ``mean $\pm$std" format.}
\label{munit}
\vspace{-3mm}
\footnotesize
\begin{center}
\begin{tabular}{l | r | r | r r}
    \multicolumn{2}{c|}{} & \multicolumn{1}{c|}{\textbf{Learning Bias}} & \multicolumn{2}{c}{\textbf{Design Bias}} \\ 
     \hline
    \multirow{2}{*}{\textbf{Metric}} & Original & w/o Latent & w/o & w/o Instance\\
    & Model & Regression (LR) & AdaIN & Normalization (IN)\\ 
        \hline
        \multicolumn{1}{l|}{$DC(C,\underline{s})$ ($\downarrow$)}   & 0.44 $\pm 0.06$ & \textbf{0.40} $\pm 0.08$ & 0.43 $\pm 0.01$ & 0.66 $\pm 0.03$\\
        \multicolumn{1}{l|}{ $DC(I,C)$  ($\uparrow$)}     & 0.57 $\pm 0.07$ & 0.57 $\pm 0.08$ & 0.58 $\pm 0.08$ & \textbf{0.73} $\pm 0.03$\\ 
        \multicolumn{1}{l|}{$DC(I,\underline{s})$  ($\uparrow$)}     & 0.70 $\pm 0.02$ & \textbf{0.73} $\pm 0.03$ & 0.56 $\pm 0.03$ & 0.63 $\pm 0.05$\\ 
        \multicolumn{1}{l|}{$IOB(I,C)$  ($\uparrow$)}    & 4.36 $\pm 0.38$ & 4.34 $\pm 0.58$ & 4.85 $\pm 0.10$ & \textbf{5.01} $\pm 0.12$ \\ 
        \multicolumn{1}{l|}{$IOB(I,\underline{s})$  ($\uparrow$)}    & 1.31 $\pm 0.04$ & \textbf{1.46} $\pm 0.05$ & 1.17 $\pm 0.04$& 1.28 $\pm 0.06$\\
        \hline
        \multicolumn{1}{l|}{$FID$ ($\downarrow$)}       & 73.48 $\pm 8.35$ & 104.51 $\pm 4.21$ & \textbf{52.48} $\pm 5.03$& 71.4 $\pm 4.86$ \\
        \multicolumn{1}{l|}{$LPIPS$ ($\uparrow$)}  & 0.08 $\pm 0.01$ & 0.09 $\pm 0.01$ & 0.06 $\pm 0.01$ & \textbf{0.10} $\pm 0.01$ \\
        \hline
\end{tabular}
\end{center}
\vspace{-4mm}
\end{table*}

\subsection{Image-to-Image Translation} 
We consider the original MUNIT and three variants:
\textbf{i)} we replace the AdaIN modules of the decoder with simple style concatenations, reducing the restrictions on the re-combination of C and S. 
\textbf{ii)} We remove the LR loss, responsible for the style following a Gaussian.
\textbf{iii)} We remove IN from the content encoder, to confirm that it helps to cancel out original style and retain the content only~\cite{adain2017iccv}.
As~\cite{munit} we evaluate quality and diversity of the translated images using the Fr\'{e}chet Inception Distance ($FID$)~\cite{heusel2017gans} and LPIPS~\cite{zhang2018cvpr}.

\textbf{Results.}
Table~\ref{munit} reports the results of the ablations on the SYNTHIA~\cite{synthia} and City-scapes~\cite{cityscapes} datasets. Replacing AdaIN (\textbf{w/o AdaIN}) with simple concatenation does not affect the level of C-S disentanglement, but it leads to a 0.14 absolute decrease in $IOB(I,\underline{s})$ and $DC(I,\underline{s})$, indicating that the style becomes less informative and less correlated with the input. Here, we observe an information shift to the content (lower $IOB(I,\underline{s})$, higher $IOB(I,C)$) leading to better translation quality but worse diversity ($LPIPS=0.06$). We infer that this variant is worse than the original model, which had more balanced quality/diversity scores.
By removing the LR learning bias (\textbf{w/o LR}), the style becomes more correlated to the input image. If the style distribution is no longer Gaussian, the style has more degrees of freedom to encode non-relevant information, which contributes to higher $IOB(I,\underline{s})$ and higher C-S disentanglement. This ablation leads to a significant translation quality decrease, while contrary to the analysis in~\cite{munit}, the diversity is not negatively affected.
Finally, by removing IN (\textbf{w/o IN}) we expect a more entangled content that is encoding also some style information. Our expectations are confirmed by the decrease in C-S disentanglement ($DC(C,S) = 0.66$), and a more informative content (which is also more correlated to the input image). Interestingly, relaxing the content constraints for a task that does not require a strictly semantic content (such as image segmentation), leads to the best quality/diversity balance. Note that we define the best balance as achieving the highest average ranking in $FID$ and $LPIPS$ (\textit{e.g.}~the ``w/o IN" model variant is the $1^{\mathrm{st}}$ in $LPIPS$ and $2^{\mathrm{nd}}$ in $FID$).

\textbf{Summary.}
Our experiments reveal a trade-off between the translation quality/diversity and disentanglement in a translation task.\footnote{Note that the effect of C-S disentanglement on task performance also depends on the data bias. An indicative example is the ``edges-to-shoes" setting where the translation is between zero-style and normal images.} Our metrics indicate that a partially disentangled C-S space --with a near-Gaussian style latent space-- leads to the best quality/diversity performance. For MUNIT this is achieved by removing the IN design bias.

\subsection{Medical Segmentation} 
In SDNet, content binarization and style Gaussianity
are the key representation constraints. We evaluate their effect and those of decoder design on segmentation performance measuring the Dice Score~\cite{dice1945measures, sorensen1948method} after: 
\textbf{i)} removing content thresholding (w/o Binarization),
\textbf{ii)} removing style Gaussianity (w/o Kullback-Liebler Divergence (KLD) and LR), 
and \textbf{iii)} considering a new decoder, obtained replacing the FiLM style conditioning with \emph{SPADE}~\cite{park2019semantic}. SPADE is less restrictive, allowing the style to encode more image-related information, such as textures, rather than just intensity (see Appendix F.1).

\begin{table*}[t]
\caption{Comparative evaluation of SDNet variants using the proposed metrics. We use the $Dice$ score to measure semantic segmentation performance on the ACDC~\cite{acdc} dataset with $1.5\%$ annotation masks. Results are in ``mean $\pm$std" format.}
\label{tbl:sdnet2}
\vspace{-2mm}
\footnotesize
\begin{center}
\begin{tabular}{l | r | r | r r}
     \multicolumn{2}{c|}{} & \multicolumn{1}{c|}{\textbf{Learning Bias}} & \multicolumn{2}{c}{\textbf{Design Bias}} \\ 
     
    \hline
    \multirow{ 2}{*}{\textbf{Metric}} & Original & w/o KLD & w/o & \multirow{ 2}{*}{SPADE} \\
    & Model & and Latent Reg. (LR) & Binarization &\\
    \hline
    $DC(C,\underline{s})$  ($\downarrow$)  & 0.49 $\pm$0.02 & 0.64 $\pm$0.03 & \textbf{0.44} $\pm$0.00 & 0.52 $\pm$0.01\\
    $DC(I,C)$  ($\uparrow$)    & 0.94 $\pm$0.01 & 0.94 $\pm$0.01 & \textbf{0.98} $\pm$0.02 & 0.93 $\pm$0.01\\
    $DC(I,\underline{s})$  ($\uparrow$)    & 0.43 $\pm$0.02 & \textbf{0.66} $\pm$0.00 & 0.44 $\pm$0.01 & 0.45 $\pm$0.01\\
    $IOB(I,C)$  ($\uparrow$)   & 4.71 $\pm$0.26 & 4.84 $\pm$0.23 & \textbf{5.89} $\pm$0.22 & 5.09 $\pm$0.00\\ 
    $IOB(I,\underline{s})$  ($\uparrow$)   & 1.00 $\pm$0.01 & 1.00 $\pm$0.04 & 0.98 $\pm$0.04 & 1.00 $\pm$0.04\\ 
    \hline
    $Dice$ ($\uparrow$) & 0.62 $\pm$0.02 & 0.61 $\pm$0.04 & 0.63  $\pm$0.04 &   \textbf{0.75} $\pm$0.02\\ 
    \hline
\end{tabular}
\end{center}
\vspace{-4mm}
\end{table*}

\textbf{Results.}
Table~\ref{tbl:sdnet2} reports our findings on the ACDC~\cite{acdc} dataset. We highlight that when using all the available annotations (fully supervised learning), all SDNet variants achieve a similar accuracy (see Appendix F.2 for more details), suggesting that strong learning biases, such as supervised segmentation costs, make disentanglement less important. Thus, we consider the semi-supervised training case with minimal supervision, using only the 1.5\% of available labelled data.
Overall, the style encodes little information in all SDNet variants, probably because all medical images in ACDC have similar styles (data bias), and reconstructing using an average style is enough to have low $IOB(I, S)$. 
However, C-S disentanglement is still important to obtain a good content representation. For example, intermediate levels of disentanglement (\textbf{SPADE}) lead to the best segmentation performance. In this variant,  disentanglement decreases compared to the original model, as some style information is probably leaked to the content (higher $DC(C,\underline{s})$ and $IOB(I,C)$). On the other hand, also removing C binarization (\textbf{w/o Binarization}) makes content more informative; since the correlation between C and S decreases, we assume that the extra information encoded in C is not part of the style.
Lastly, removing the Gaussian prior constraints from the style (\textbf{w/o KLD and LR}) leads to the lowest degree of disentanglement as there is no information bottleneck on S, and a slight decrease of the Dice score.

\textbf{Summary.}
We find disentanglement to have minimal effect on task performance when training with strong learning signals (\textit{i.e.}~supervised costs). 
In the semi-supervised setting, a higher (but not full) degree of disentanglement leads to better performance, while the amount of information in C alone is not enough to achieve adequate segmentation performance.

\subsection{Pose Estimation} 
We consider the original PANet model and four possible variants, relaxing design biases on both C and style, and learning biases. In detail: 
\textbf{i)} we experiment with a different conditioning mechanism to re-entangle S and C, that consists of the use of AdaIN, rather than just multiplying each S vector with a separate C channel (introducing a bias on S, similar to MUNIT). 
\textbf{ii)} We consider the case where, instead of learning a different S for each channel of C, we extract a global S vector, predicted by an MLP (relaxing the tight 1:1 correspondence between C and S channels).
\textbf{iii)} We also consider the case where each C part is not approximated by a Gaussian prior. Since we cannot use the original decoder to combine C and S, we reintroduce S using AdaIN.
\textbf{iv)} Finally, we evaluated the effect of the equivariance constraint, by removing it from the cost function.

\begin{table*}[t]
\caption{
Comparative evaluation of PANet variants using the proposed metrics. We use $SIM$ to measure the performance in terms of pose estimation from landmarks on the DeepFashion~\cite{liuLQWTcvpr16DeepFashion} dataset. Results are in ``mean $\pm$std" format.
}
\label{tab:posenet}
\vspace{-3mm}
\footnotesize
\begin{center}
\begin{tabular}{l | r | r | r r r }
    \multicolumn{2}{c|}{} & \multicolumn{1}{c|}{\textbf{Learning Bias}} & \multicolumn{3}{c}{\textbf{Design Bias}} \\ \hline
    \multirow{ 2}{*}{\textbf{Metric}} & Original & \multirow{ 2}{*}{w/o Equivar.} & AdaIN & \multirow{ 2}{*}{AdaIN} & \multirow{ 2}{*}{MLP}\\
    & Model & & w/o Gaussian & &\\ \hline
    $DC(C,\underline{s})$  ($\downarrow$)  & 0.65 $\pm$0.01 & 0.76 $\pm$0.08 & \textbf{0.25} $\pm$0.01 & 0.36 $\pm$0.02 & 0.69	$\pm$0.03\\ 
    $DC(I,C)$  ($\uparrow$)    & 0.59 $\pm$0.01 &  \textbf{0.60}  $\pm$0.02 & 0.53 $\pm$0.01 & 0.56 $\pm$0.01 & 0.58 $\pm$0.02 \\ 
    $DC(I,\underline{s})$  ($\uparrow$)    & \textbf{0.83} $\pm$0.01 & 0.82  $\pm$0.01 & 0.38 $\pm$0.06& 0.81 $\pm$0.01 & 0.82 $\pm$0.03\\ 
    $IOB(I,C)$  ($\uparrow$) & 1.50 $\pm$0.08 & 1.50  $\pm$0.08 & \textbf{1.53} $\pm$0.06& 1.52 $\pm$0.08 & 1.49 $\pm$0.06\\ 
    $IOB(I,\underline{s})$  ($\uparrow$) & 1.09 $\pm$0.04 & 1.13  $\pm$0.06 & 1.12 $\pm$0.09 & 1.10 $\pm$0.15 & \textbf{1.21} $\pm$0.09 \\
    \hline
    $SIM$ ($\uparrow$) & \textbf{0.71} $\pm$0.02 & 0.47 $\pm$0.04 & 0.58 $\pm$0.00& 0.64 $\pm$0.01 & 0.68 $\pm$0.01\\
\hline
\end{tabular}
\end{center}
\vspace{-8mm}
\end{table*}

\textbf{Results.}
Table \ref{tab:posenet} reports results of the ablations on the DeepFashion~\cite{liuLQWTcvpr16DeepFashion} dataset. We assess model performance using $SIM$~\cite{bylinskii2019pami} to measure the similarity between the predicted and ground truth landmarks visualized as heatmaps.
Whilst the original model is the best to predict landmarks, it only achieves average disentanglement (see $DC(C,\underline{s})$). Using an \textbf{AdaIN}-based decoder consistently improves disentanglement as it has a strong inductive bias on the re-entangled representation (see $DC(C,\underline{s})$ for AdaIN, and \textbf{AdaIN w/o Gaussian}), but it leads to worse landmark detection -- the representation adapts tightly to the strongly-biased decoder, and the content loses transferability to other tasks, and interpretability (see Figs.~\ref{fig:qual},~\ref{fig:supp_ppnet}).
Using an \textbf{MLP} to encode S relaxes the specific conditioning between C and S (a design bias) and reduces disentanglement. In fact, there is an information shift from C to S, as indicated by the higher $IOB(I,\underline{s})$, and we observe a high $DC(C,\underline{s})$. Here, a moderate decrease of disentanglement shows slightly lower task performance.
Finally, the equivariance cost is the most important factor for disentanglement; removing it (\textbf{w/o Equivariance}) leads to the most entangled representation (high $DC(C,\underline{s})$), and accuracy decrease in landmark detection.

\textbf{Summary.}
Overall, lowering disentanglement leads to better landmark detection. Again, balance is the key to improve the auxiliary tasks. In PANet, partial disentanglement is achieved by carefully balancing the design biases used to extract the style and to reintroduce it to the content while decoding. Relaxing such biases with AdaIN or MLP makes landmark detection worse.

\subsection{Discussion}
We now discuss the relationship between C-S disentanglement and inductive biases, task performance, interpretability of the latent representations.

\textbf{Do biases affect C-S disentanglement?} 
Results in Sec.~\ref{experiments} illustrate that learning and design biases critically affect disentanglement. However, no evaluation can specifically characterize the relative importance of each one, since this depends on the task at hand, as well as the utilized data.
In MUNIT, disentanglement is mainly encouraged by the content-related design and learning biases. In fact, IN is key to removing style information from the content, and the model cannot be successfully trained without LR of the content.
Disentanglement in SDNet is susceptible to the biases that affect both latent variables. Using a SPADE decoder or removing content thresholding leads to more entanglement, while making the style Gaussian through learning constraints restricts its informativeness and encourages disentanglement.
Similarly, PANet disentanglement is affected both by designing the content as Gaussian, and by the equivariance of C and S \textit{w.r.t.}~spatial or intensity transformations, respectively.

\begin{figure*}
    \centering
    \includegraphics[width=\textwidth]{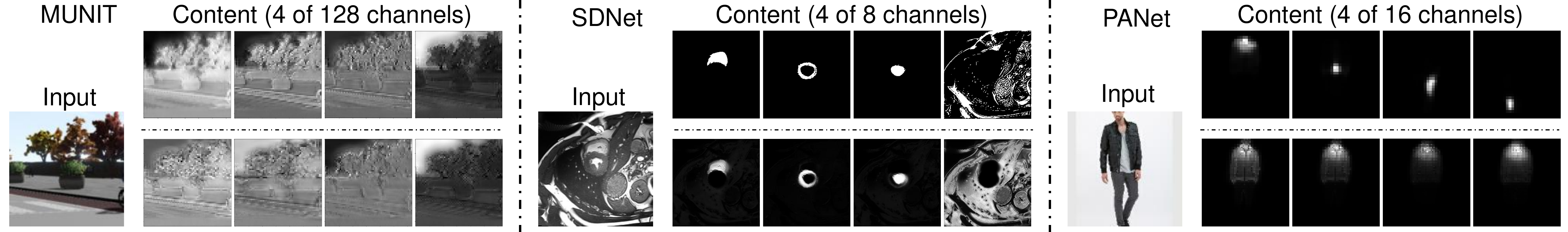}
    \vspace{-0.2in}
    \caption{
    Content \textit{interpretability} of each original model (top row) and a variant with the most correlated C and S  (bottom row). Removing content-related design biases from SDNet and PANet leads to less interpretable representations (same objects/joints appear in different channels), such as the unconstrained content representations of MUNIT.}
    \label{fig:qual}
    \vspace{-2mm}
\end{figure*}

\textbf{What is the relationship between C-S disentanglement and task performance?} Our results showcase a clear sweet spot between C-S disentanglement and downstream task performance. In particular, we observe that lowering disentanglement by relaxing constraints on the content (\textit{e.g.}~removing IN), but preserving the biases that enforce style priors, such as C-S equivariance, leads to better performance.

\textbf{Does disentanglement affect content interpretability?}
Interpretability is hard to quantify without metrics. Here, we consider the C interpretable if distinct objects appear in different channels. We qualitatively analyze C interpretability in Fig.~\ref{fig:qual} (see also Appendix I).
Interpretability varies a lot with different \textit{design biases} of the model, while learning biases do not seem to affect it. Without restrictive design bottlenecks on C,
MUNIT spreads the content across channels. Instead, SDNet and PANet original models encourage C to encode different objects, or parts, into different channels.
In SDNet, a semantic content is encouraged by applying Softmax across channels and then binarize the output features, while PANet approximates body parts as 2D Gaussians enforcing an information bottleneck on each channel of C.
Removing the C constraints from SDNet and PANet spreads the spatial information across all channels, decreasing interpretability.

\section{Conclusion}
\label{conclusion}
In this paper we evaluated the disentanglement between image C and S through experimenting on 3 SoTA models, and showcased how design and learning biases affect disentanglement and by extension task performance. Our findings suggest that whilst content-style disentanglement enables the implementation of certain equivariant tasks, partially (dis)entangled can lead to better performance than fully disentangled ones.
Additionally, our analysis suggests that strict design constraints on the content space lead to increased interpretability, which could be exploited in post-hoc tasks. Using our findings and the presented metrics will enable the design of better models that achieve the degree of disentanglement that maximizes performance, rather than blindly pursuing very high (or low) disentanglement.

\bibliography{egbib}

\newpage
\appendix

\section{Maximizing Likelihood by Minimizing Mean Square Error}
\label{app:iob}
Let $y$ denote a generic pixel in an image $I$, and $\tilde{y}$ the respective pixel in the reconstructed image $\tilde{I}$, obtained trough a learned decoding function.

If we assume the reconstruction error, denoted as $\varepsilon$, to be normally distributed (\textit{i.e.}~$\varepsilon \sim \mathcal{N}\left(0, \sigma^{2}\right)$), 
then, the predicted value $\tilde{y}$ is normally distributed around the true value $y$, thus $\tilde{y} \sim \mathcal{N}\left(y, \sigma^{2}\right)$.
Based on this assumption, the probability density function can be defined as:
\begin{equation}
    f\left(\tilde{y}|y, \sigma^{2}\right)=\frac{1}{\sqrt{2 \pi \sigma^{2}}} e^{-\frac{(\tilde{y}-y)^{2}}{2 \sigma^{2}}} .
\end{equation}
Given a set of observations, \textit{e.g.}~the pixels of the image, we maximize the likelihood $\mathcal{L}$ as the product of the probability densities of the observations:
\begin{equation}
    \mathcal{L} 
    = \prod_{i=1}^{n} f\left(y_{i} | \tilde{y}_{i}, \sigma^{2}\right)
    =\left(2 \pi \sigma^{2}\right)^{-n / 2} e^{-\frac{\sum_{i=1}^{n}\left(y_{i}-\tilde{y}_{i}\right)^{2}}{2 \sigma^{2}}} .
\end{equation}
Assuming the variance of the error to be independent from the input variables, optimizing the latter formula is equivalent to optimize:
\begin{equation}
    \log \bigg( \frac{\mathcal{L}}{\left(2 \pi \sigma^{2}\right)^{-n / 2}} \bigg) = -\frac{\sum_{i=1}^{n}\left(y_{i}-\tilde{y}_{i}\right)^{2}}{2 \sigma^{2}} .
\end{equation}
Thus, maximizing the original likelihood function is equivalent to minimizing $\sum_{i=1}^{n}\left(y_{i}-\tilde{y}_{i}\right)^{2}$ that is the scaled Mean Squared Error (MSE). Thus, by training the decoder to minimize MSE, we train it to maximize the Mutual Information (MI) between $z$ and $I$.

After training the decoder $G_{\theta}$ (see Sec.~3), computing MSE equivalent to directly measuring the MI. There is a relationship between likelihood and MSE (shown below), but the likelihood acts as a lower bound to MI.

\textbf{Relationship MSE - likelihood:}
Note that if we divide both parts of the equation by $n$ and then we multiply by $- 2\sigma^2$, we obtain:

\begin{equation}
   \sum_{i=1}^{n}\frac{\left(y_{i}-\tilde{y}_{i}\right)^{2}}{n} = 
    - \frac{2\sigma^2}{n}\cdot \log \frac{\mathcal{L}}{(2 \pi \sigma^2)^{-n/2}},
\end{equation}
that is:
\begin{equation}
\label{eq_mse}
   MSE = - \frac{2\sigma^2}{n} \log (\mathcal{L}) - \sigma^2\log({2 \pi \sigma^2}).
\end{equation}
Since we assume homoscedastic distributions, \textit{i.e.}~fixed $\sigma^2$, Equation~\ref{eq_mse} can be expressed as:
\begin{equation}
   MSE =  - \frac{a}{n} \log (\mathcal{L}) - b,
\end{equation}
where $a$ and $b$ are positive constants.

\section{Empirical Study with the Teapot Dataset}

\begin{figure*}[t]
    \centering
    \includegraphics[width=1\textwidth]{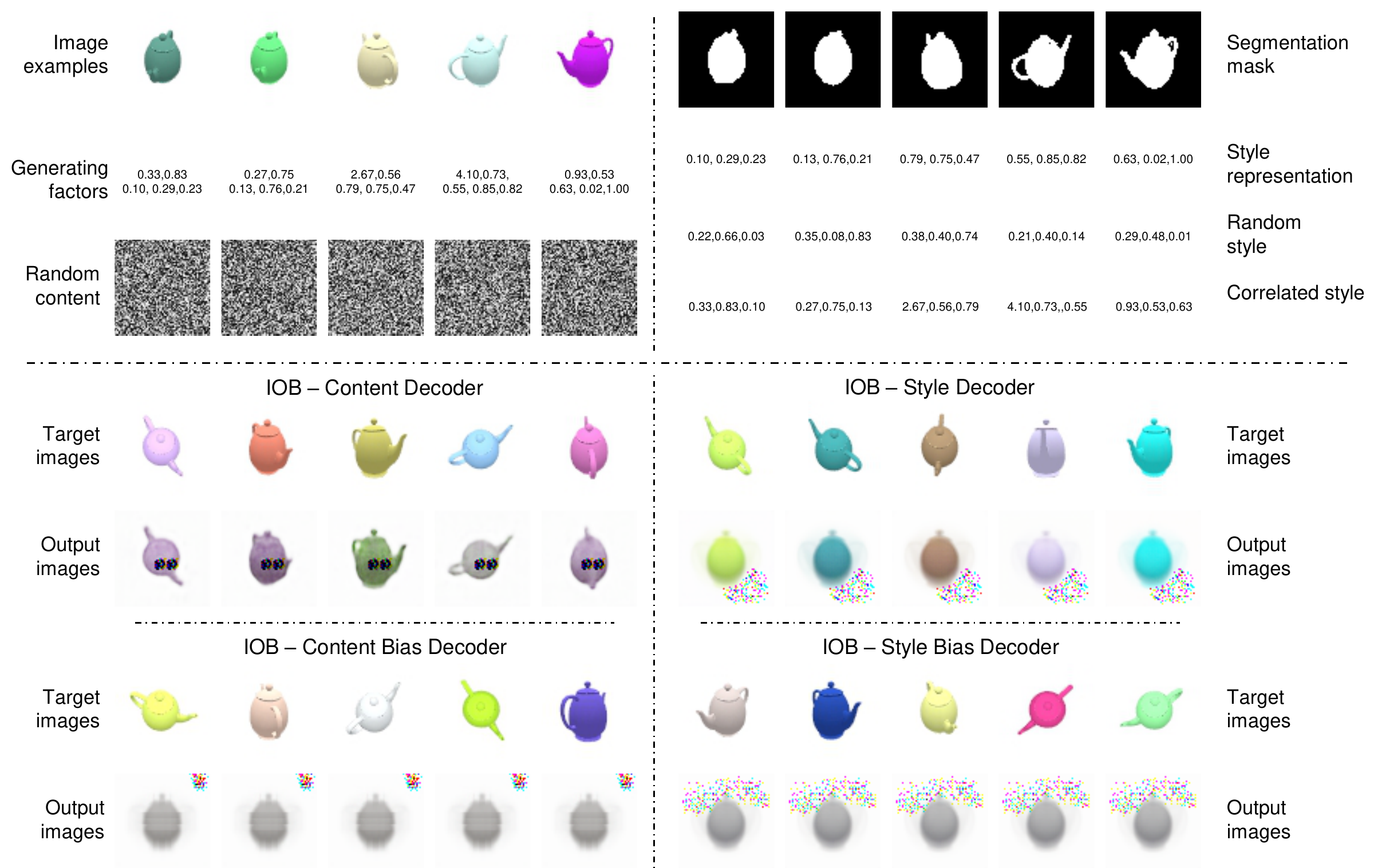}
    \caption{Visuals for the empirical study with the teapot dataset. Top: examples of original images, ground truth generating factors and segmentation masks. We also show the randomly sampled content and style representations. Bottom: examples of target images and output images for the $IOB$ decoders.}
    \label{fig:supp_teapot}
\end{figure*}

Visual examples and qualitative results of the empirical study on the proposed metrics with the teapot dataset are included in Fig. \ref{fig:supp_teapot}. It is notable that the artifacts in the reconstructed images introduced by the decoder bias are observed in the results of both decoders and bias decoders. 

\begin{table}[t]
\caption{$IOB$ decoders design for the teapot dataset.}
\label{tbl:IOBTeapot}
\vspace{-0.1in}
\footnotesize
\begin{center}
\begin{tabular}{c|r@{$\to$}l|c}
        \hline
        Decoder &  Input Shape & Output Shape   & Layer Information\\
        \hline
        \multirow{8}*{$G_\theta(C)$} & (1,64,64)&(8,64,64)   &   CONV-(O:8,K:7x7,S:1,P:3), IN, Leaky ReLU\\
        ~ & (8,64,64)   &(16,32,32)	    &   CONV-(O:16,K:4x4,S:2,P:1), IN, Leaky ReLU\\
        ~ & (16,32,32)  & (32,16,16)	&   CONV-(O:32,K:4x4,S:2,P:1), IN, Leaky ReLU\\
        ~ & (32,16,16)  & (64,8,8)	    &   CONV-(O:64,K:4x4,S:2,P:1), IN, Leaky ReLU\\
        ~ & (64,8,8)    & (32,16,16)	&   DECONV-(O:32,K:4x4,S:2,P:1), IN, Leaky ReLU\\
        ~ & (32,16,16)  & (16,32,32)	&   DECONV-(O:16,K:4x4,S:2,P:1), IN, Leaky ReLU\\
        ~ & (16,32,32)  & (8,64,64)	    &   DECONV-(O:8,K:4x4,S:2,P:1),  IN, Leaky ReLU\\
        ~ & (8,64,64)   & (3,64,64)	    &   CONV-(O:3,K:7x7,S:1,P:3),  Tanh\\
        \hline
        \multirow{6}*{$G_\theta(\underline{s})$} & (3)&(256)   &   FC-(O:256)\\
        ~ & (256)    &(4096)	&   FC-(O:4096), Flatten\\
        ~ & (64,8,8)    & (32,16,16)	&   DECONV-(O:32,K:4x4,S:2,P:1), IN, Leaky ReLU\\
        ~ & (32,16,16)  & (16,32,32)	&   DECONV-(O:16,K:4x4,S:2,P:1), IN, Leaky ReLU\\
        ~ & (16,32,32)  & (8,64,64)	    &   DECONV-(O:8,K:4x4,S:2,P:1),  IN, Leaky ReLU\\
        ~ & (8,64,64)   & (3,64,64)	    &   CONV-(O:3,K:7x7,S:1,P:3),  Tanh\\
        \hline
\end{tabular}
\end{center}
\vspace{-0.1in}
\end{table}

\begin{table}[t]
\caption{$IOB$ decoders design for MUNIT.}
\label{tbl:IOBMUNIT}
\vspace{-0.1in}
\footnotesize
\begin{center}
\begin{tabular}{c|r@{$\to$}l|c}
        \hline
        Decoder &  Input Shape & Output Shape   & Layer Information\\
        \hline
        \multirow{7}*{$G_\theta(C)$} & (128,64,64)&(128,64,64)   &   CONV-(O:128,K:7x7,S:1,P:3), IN, Leaky ReLU\\
        ~ & (128,64,64) & (128,32,32)	&   CONV-(O:128,K:4x4,S:2,P:1), IN, Leaky ReLU\\
        ~ & (128,32,32) & (128,16,16)	&   CONV-(O:128,K:4x4,S:2,P:1), IN, Leaky ReLU\\
        ~ & (128,16,16) & (64,32,32)	&   DECONV-(O:64,K:4x4,S:2,P:1), IN, Leaky ReLU\\
        ~ & (64,32,32)  & (32,64,64)	&   DECONV-(O:32,K:4x4,S:2,P:1), IN, Leaky ReLU\\
        ~ & (32,64,64)  & (16,128,128)  &   DECONV-(O:16,K:4x4,S:2,P:1),  IN, Leaky ReLU\\
        ~ & (16,128,128)& (3,128,128)	&   CONV-(O:3,K:7x7,S:1,P:3),  Tanh\\
        \hline
        \multirow{8}*{$G_\theta(\underline{s})$} & (8)&(256)   &   FC-(O:256)\\
        ~ & (256)    &(4096)	&   FC-(O:4096)\\
        ~ & (4096)   &(8192)	&   FC-(O:8192), Flatten\\
        ~ & (128,8,8)   & (64,16,16)	&   DECONV-(O:64,K:4x4,S:2,P:1), IN, Leaky ReLU\\
        ~ & (64,16,16)  & (32,32,32)	&   DECONV-(O:32,K:4x4,S:2,P:1), IN, Leaky ReLU\\
        ~ & (32,32,32)  & (16,64,64)	&   DECONV-(O:16,K:4x4,S:2,P:1),  IN, Leaky ReLU\\
        ~ & (16,64,64)  & (8,128,128)	&   DECONV-(O:8,K:4x4,S:2,P:1),  IN, Leaky ReLU\\
        ~ & (8,128,128) & (3,128,128)	&   CONV-(O:3,K:7x7,S:1,P:3),  Tanh\\
        \hline
\end{tabular}
\end{center}
\vspace{-0.1in}
\end{table}

\begin{table}[t]
\caption{$IOB$ decoders design for SDNet.}
\label{tbl:IOBSDNet}
\vspace{-0.1in}
\footnotesize
\begin{center}
\begin{tabular}{c|r@{$\to$}l|c}
        \hline
        Decoder &  Input Shape & Output Shape   & Layer Information\\
        \hline
        \multirow{10}*{$G_\theta(C)$} & (8,224,224)&(8,224,224)   &   CONV-(O:8,K:7x7,S:1,P:3), IN, Leaky ReLU\\
        ~ & (8,224,224) & (16,112,112)  &   CONV-(O:16,K:4x4,S:2,P:1), IN, Leaky ReLU\\
        ~ & (16,112,112)& (32,56,56)	&   CONV-(O:32,K:4x4,S:2,P:1), IN, Leaky ReLU\\
        ~ & (32,56,56)  & (64,28,28)	&   CONV-(O:64,K:4x4,S:2,P:1), IN, Leaky ReLU\\
        ~ & (64,28,28)  & (128,14,14)	&   CONV-(O:128,K:4x4,S:2,P:1), IN, Leaky ReLU\\
        ~ & (128,14,14) & (64,28,28)	&   DECONV-(O:64,K:4x4,S:2,P:1), IN, Leaky ReLU\\
        ~ & (64,28,28)  & (32,56,56)	&   DECONV-(O:32,K:4x4,S:2,P:1), IN, Leaky ReLU\\
        ~ & (32,56,56)  & (16,112,112)	&   DECONV-(O:16,K:4x4,S:2,P:1), IN, Leaky ReLU\\
        ~ & (16,112,112)& (8,224,224)	&   DECONV-(O:8,K:4x4,S:2,P:1),  IN, Leaky ReLU\\
        ~ & (8,224,224) & (1,224,224)   &   CONV-(O:1,K:7x7,S:1,P:3),  Tanh\\
        \hline
        \multirow{8}*{$G_\theta(\underline{s})$} & (3)&(256)   &   FC-(O:256)\\
        ~ & (256)    &(4096)	&   FC-(O:4096)\\
        ~ & (4096)   &(25088)	&   FC-(O:25088), Flatten\\
        ~ & (128,14,14) & (64,28,28)	&   DECONV-(O:64,K:4x4,S:2,P:1), IN, Leaky ReLU\\
        ~ & (64,28,28)  & (32,56,56)	&   DECONV-(O:32,K:4x4,S:2,P:1), IN, Leaky ReLU\\
        ~ & (32,56,56)  & (16,112,112)	&   DECONV-(O:16,K:4x4,S:2,P:1), IN, Leaky ReLU\\
        ~ & (16,112,112)& (8,224,224)	&   DECONV-(O:8,K:4x4,S:2,P:1),  IN, Leaky ReLU\\
        ~ & (8,224,224) & (1,224,224)   &   CONV-(O:1,K:7x7,S:1,P:3),  Tanh\\
        \hline
\end{tabular}
\end{center}
\vspace{-0.1in}
\end{table}

\begin{table}[t]
\caption{$IOB$ decoders design for PANet.}
\label{tbl:IOBPANet}
\vspace{-0.1in}
\footnotesize
\begin{center}
\begin{tabular}{c|r@{$\to$}l|c}
        \hline
        Decoder &  Input Shape & Output Shape   & Layer Information\\
        \hline
        \multirow{5}*{$G_\theta(C)$} & (3,64,64)&(16,64,64)   &   CONV-(O:16,K:7x7,S:1,P:3), IN, Leaky ReLU\\
        ~ & (16,64,64)   & (32,32,32)  &   CONV-(O:32,K:4x4,S:2,P:1), IN, Leaky ReLU\\
        ~ & (32,32,32)  & (16,64,64)	&   DECONV-(O:16,K:4x4,S:2,P:1), IN, Leaky ReLU\\
        ~ & (16,64,64)  & (8,128,128)	&   DECONV-(O:8,K:4x4,S:2,P:1), IN, Leaky ReLU\\
        ~ & (8,128,128) & (3,128,128)   &   CONV-(O:3,K:7x7,S:1,P:3),  Tanh\\
        \hline
        \multirow{4}*{$G_\theta(\underline{s})$} & (1024)&(1,32,32)   &   Flatten\\
        ~ & (1,32,32) & (16,64,64)	&   DECONV-(O:16,K:4x4,S:2,P:1), IN, Leaky ReLU\\
        ~ & (16,64,64)  & (8,128,128)	&   DECONV-(O:8,K:4x4,S:2,P:1), IN, Leaky ReLU\\
        ~ & (8,128,128) & (3,128,128)   &   CONV-(O:3,K:7x7,S:1,P:3),  Tanh\\
        \hline
\end{tabular}
\end{center}
\vspace{-0.1in}
\end{table}

\section{Model Design and Training Scheme for $IOB$}
For the structure of $IOB$ decoders, we vary the number of layers for different applications due to the different dimensions of the representations. The design of the decoders for the teapot dataset, MUNIT, SDNet and PANet can be found in Table \ref{tbl:IOBTeapot}, \ref{tbl:IOBMUNIT}, \ref{tbl:IOBSDNet} and \ref{tbl:IOBPANet}. The notations in the tables are:
O: the number of output channels;
K: the kernel size;
S: the stride size;
P: the padding size; FC: fully-connected layer;
IN: instance normalization; Overall, $G_\theta(\underline{s})$ consists of several linear layers, followed by transpose (upsampling steps) and one plain CONV layer that generates the final image. $G_\theta(C)$ follows an autoencoder structure with several encoder and decoder CONV layers. For the teapot dataset, the content representation has size $1\times64\times64$ and the style representation has size $3$. For MUNIT, the content representation has size $128\times64\times64$ and the style representation has size $8$. For SDNet, the content representation has size $8\times224\times224$ and the style representation has size $8$. For PANet, the content representation has size $3\times64\times64$ and the style representation has size $1024$. Note that it is not necessary to have exactly same design as in the tables, where the key suggestion is to design the decoders to generate as high-quality as possible reconstructed images. 

All the decoders are trained using the Adam optimiser \cite{adam} ($\beta_1=0.5, \beta_2=0.999$) with a learning rate of $1e^{-4}$ for 40 epochs using batch size 10.

\begin{table}[t]
\caption{Overview of the \textit{design} and \textit{learning biases} that are investigated in the context of the three investigated vision tasks: a) image-to-image translation (MUNIT), b) medical segmentation (SDNet), and c) pose estimation (PANet).}
\label{tbl:biases}
\vspace{-0.1in}
\footnotesize
\begin{center}
\begin{tabular}{l | r | c | c | c}
    \multicolumn{2}{c|}{} & \textbf{MUNIT} & \textbf{SDNet} & \textbf{PANet} \\
    \hline
    \multirow{ 6}{*}{\textbf{Design Bias}} & AdaIN & $\surd$ &  & $\surd$ \\
                                  \cline{2-5}
                                  & Instance & \multirow{2}{*}{$\surd$} & \multirow{2}{*}{} & \multirow{2}{*}{}  \\
                                  & Normalization & & & \\
                                  \cline{2-5}
                                  & SPADE & & $\surd$ & \\
                                  \cline{2-5}
                                  & Binarization & & $\surd$ &  \\
                                  \cline{2-5}
                                  & MLP & & & $\surd$ \\
    \hline
    \multirow{ 4}{*}{\textbf{Learning Bias}} & Latent & \multirow{2}{*}{$\surd$} &                                          \multirow{2}{*}{$\surd$}  &   \\ 
                                  & Regression & & & \\
                                  \cline{2-5}
                                  & KL Divergence & & $\surd$ &  \\
                                  \cline{2-5}
                                  & Equivariance & &  & $\surd$ \\
    \hline
\end{tabular}
\end{center}
\vspace{-0.1in}
\end{table}

\section{Detailed Application Description}
\label{app:methods}
Table~\ref{tbl:biases} summarizes the \textit{design} and \textit{learning biases} of the methods presented in Sec.~5. Note that the biases are reported as modules, without indicating the way they are used in our experiments (\textit{e.g.}~AdaIN is reported without specifying that it is removed from the original MUNIT, but is added to PANet as a variant).

\subsection{MUNIT for Image-to-Image Translation}
Multimodal Unsupervised Image-to-image Translation (MUNIT)~\cite{munit} does not impose strict constraints on the learned representations, and achieves disentanglement with both design and learning biases. 

The basic assumption is that multi-domain images (a necessary \textit{data bias}), share common content information, but differ in style. A content encoder maps images to multi-channel feature maps, by removing style with IN layers~\cite{adain2017iccv} (\textit{design bias}). A second encoder extracts global style information with fully connected layers and global pooling. Finally, style and content are combined in a decoder with AdaIN modules~\cite{adain2017iccv} (\textit{design bias}). 

Disentanglement is additionally promoted with a bidirectional reconstruction loss~\cite{zhu2017toward} that enables style transfer. In order to learn a smooth representation manifold, two LR losses (\textit{learning bias}) are applied on content and style extracted from input images: content LR penalizes the distance to the content extracted from reconstructed images, whereas style LR encourages encoded style distributions to match their Gaussian priors. Finally, adversarial learning encourages realistic synthetic images.

\subsection{SDNet for Medical Image Segmentation}
SDNet~\cite{chartsias2019factorised} is a semi-supervised framework that disentangles medical images in anatomical features (content) and imaging-specific characteristics (style). Similarly to other models, SDNet uses separate content and style encoders, but here a segmentation network is applied on the content features trained with supervised objectives and annotated images (\textit{data bias}). 

However, in contrast to MUNIT, SDNet does not impose a design bias on the encoder, but rather on the content which is represented as multi-channel binary maps of the same resolution as the input (\textit{design bias}).

This is obtained with a softmax and a thresholding function with the straight-through operator~\cite{bengio2013estimating}, such that any style is removed from the content.
To encourage style features to encode residual information (and not content), a loss enforces the style representation to approximate a standard Gaussian, following the VAE formulation~\cite{kingma2013auto} (\textit{learning bias}). In this setup, any information encoded in style comes at a cost, and thus encoding redundant information is prevented~\cite{alemi2016deep}. Furthermore, a LR loss of the style is employed to prevent posterior collapse of the decoder (\textit{learning bias}).

Finally, style and content are combined to reconstruct the input image by applying a series of convolutional layers with feature-wise linear modulation (FiLM) conditioning. Similarly to AdaIN, FiLM modules are restrictive, allowing the style only to normalize the conditioned feature maps, and thus further discouraging the style from encoding content information (\textit{design bias}).

\subsection{PANet for Pose Estimation}
For the pose estimation task, we consider a dual-stream autoencoder denoted as PANet~\cite{lorenz2019unsupervised}. PANet consists of two branches that decouple pose (content) and appearance (style) but employs heavily entangled encoders-decoders. 

The content is represented as a multi-channel feature map, where each channel corresponds to a specific body part (since the number of parts are fixed, this imposes a strong \textit{data bias}).
A Gaussian distribution is applied to each feature map to remove any style information, whilst also preserving the spatial correspondence (\textit{design bias}).

The corresponding style information is extracted from the encoder features using average pooling (\textit{design bias}). More critically, style vectors do not correspond to global image style, since they are applied to specific content parts during decoding (\textit{design bias}).

Finally, disentanglement is encouraged with a transformation equivariance loss (\textit{learning bias}). This ensures that the spatial transformations, such as translations and rotations, affect only the content, while the intensity ones, such as the color and texture information, affect only the style.

\section{SYNTHIA-Cityscapes Description and MUNIT Training Setup}
\label{sec:munit_supp}
\textbf{Data.}
We use SYNTHIA~\cite{synthia}, which consists of over $20,000$ rendered images and corresponding pixel-level semantic annotations, where 13 classes of objects are labeled for aiding segmentation and scene understanding problems.
We also use Cityscapes~\cite{cityscapes}, which contains a set of diverse street scene stereo video sequences and over 5k frames of high-quality semantic annotations, where 30 classes of instances are labeled in the segmentation masks.

\textbf{Training setup.}
MUNIT achieves unsupervised multi-modal image-to-image translation by minimizing the following loss function:
\begin{equation}
    \mathcal{L}_{total} = \mathcal{L}_{GAN} + \lambda_{1}\mathcal{L}_{rec} + \lambda_{2}\mathcal{L}_{c-rec} + \lambda_{3}\mathcal{L}_{s-rec},
\end{equation}
where $\mathcal{L}_{rec}$ is the image reconstruction loss, $\mathcal{L}_{c-rec}$ and $\mathcal{L}_{s-rec}$ are the content and style reconstruction losses, and $\lambda_{1}=10$, $\lambda_{2}=1$ are the hyperparameters used by the authors in~\cite{munit}.\\

\section{ACDC Description and SDNet Training Setup}
\label{sec:sdnet_supp}
\textbf{Data.} 
We use data from the Automatic Cardiac Diagnosis Challenge (ACDC)~\cite{acdc}, which contains cardiac cine-MR images acquired from different MR scanners and resolution on 100 patients. Images were resampled to 1.37 $mm^2$/pixel resolution and cropped to $224\times 224$ pixels. Manual segmentations are provided for the left ventricular cavity, the myocardium and right ventricle in the end-systolic and end-diastolic cardiac phases. In total there are 1920 images with manual segmentations and 23,530 images with no segmentations.

\textbf{Training setup.}
SDNet is trained by minimizing the following loss function: 
\begin{equation}
    \mathcal{L}_{total} = \lambda_{1} \mathcal{L}_{KL} + \lambda_{2} \mathcal{L}_{seg} + \lambda_{3} \mathcal{L}_{rec} + \lambda_{4} \mathcal{L}_{z_{rec}},
\end{equation}
where $\mathcal{L}_{KL}$ is the KL Divergence measured between the sampled and the predicted style vectors, $\mathcal{L}_{rec}$ is the image reconstruction loss, $\mathcal{L}_{seg}$ is the anatomy segmentation loss, and $\mathcal{L}_{z_{rec}}$ is the LR loss between the sampled and the re-encoded style vector. $\lambda_{1}=0.01, \lambda_{2}=10, \lambda_{3}=1,$ and $\lambda_{4}=1$ are the hyperparameters used by the authors in~\cite{chartsias2019factorised}.

\begin{figure}[t]
    \centering
    \includegraphics[width=0.35\textwidth]{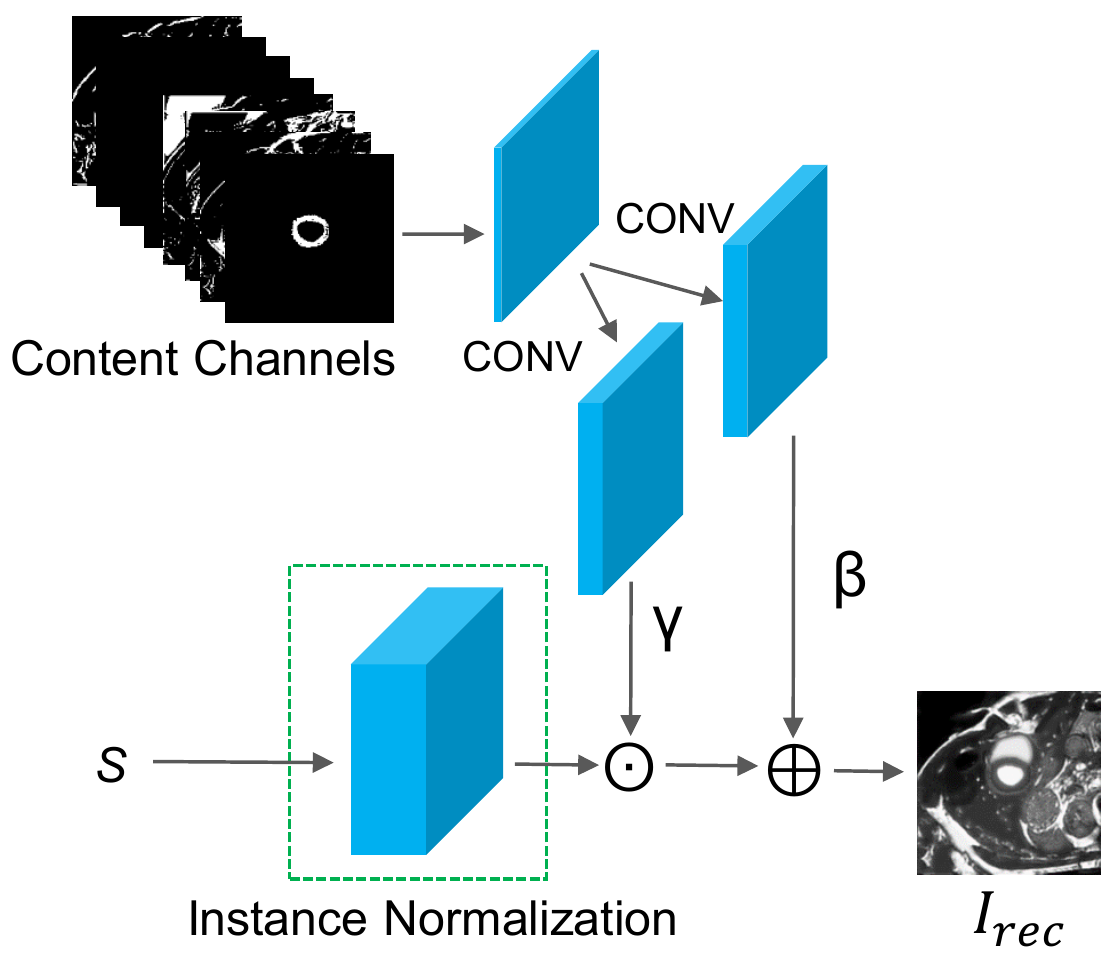}
    \caption{Detailed representation of the SPADE decoder~\cite{park2019semantic} in the context of SDNET~\cite{chartsias2019factorised}. Style is denoted as $S$, while CONV represents the convolution operation. Note that $\gamma$ and $\beta$ parameters are applied to the normalized style activations through element-wise multiplication and addition, respectively.}
    \label{fig:spade_fig}
\end{figure}

\subsection{SPADE Decoder}
\label{spade}
As described in Sec.~5, SDNet relies on a FiLM-based decoder to combine the content and style information and reconstruct the input image. The key characteristic of FiLM is that it gradually adds style information to the content-based reconstruction process. Additionally, an alternative approach for combining the content and style features is investigated, by using a SPADE decoder~\cite{park2019semantic} to further expose the design bias added by the decoder architecture.  

A SPADE block receives the content channels and projects them onto an embedding space using two convolutional layers to produce the modulation parameters (tensors) $\gamma$ and $\beta$. These parameters are then used to scale ($\gamma$) and shift ($\beta$) the normalized activations of the style representation. We utilize multiple SPADE blocks
to fuse content and style information at different levels of granularity during decoding. A schematic of the utilized SPADE decoder in the context of SDNet is depicted in Fig.~\ref{fig:spade_fig}.

\begin{table*}[b]
\caption{Comparative evaluation of SDNet~\cite{chartsias2019factorised} variants on the ACDC~\cite{acdc} dataset with $100\%$ annotation masks, using the proposed metrics. The $Dice$ $Score$ metric is used to measure the performance in terms of semantic segmentation.}
\label{tbl:sdnet100}
\begin{center}
\vspace{-0.1in}
\small
\begin{tabular}{l | r | r | r r}
     \multicolumn{2}{c|}{} & \multicolumn{1}{c|}{\textbf{Learning Bias}} & \multicolumn{2}{c}{\textbf{Design Bias}} \\ 
    \hline
    \multirow{ 2}{*}{\textbf{SDNet}} & Original & w/o KLD & w/o & \multirow{ 2}{*}{SPADE} \\
    & Model & and Latent Regression & Binarization & \\
    \hline
    $DC(C,$ $S)$ ($\downarrow$)  & 0.48 & 0.57 & 
                              \textbf{0.43} & 0.59 \\
    $DC(I,$ $C)$ ($\uparrow$)    & 0.97 & 0.95 & 
                              \textbf{0.97} & 0.94 \\
    $DC(I,$ $S)$ ($\uparrow$)    & 0.44 & 0.53 & 0.44 &                                  \textbf{0.57} \\
    $IOB(I,$ $C)$ ($\uparrow$)      & 5.66 & 3.86 &
                              \textbf{6.21} & 5.63 \\ 
    $IOB(I,$ $S)$ ($\uparrow$)      & 0.99 & 0.96  & 1.00 &                                     \textbf{1.02} \\ 
    \hline
    $Dice$ $Score$ ($\uparrow$) & 0.82 & 0.81 &
                              0.82 & \textbf{0.83}\\ 
    \hline
\end{tabular}
\end{center}
\end{table*}

\subsection{Medical Segmentation (100\% Annotations)} 
\label{sdnet_100}
In Sec.~5.2, we present the results of the SDNet model variants trained with minimal supervision, using only the 1.5\% of the provided ACDC~\cite{acdc} annotations. Here, we provide the results for the same experiment but using the 100\% of the provided annotations. From the results reported in Table~\ref{tbl:sdnet100}, it can be seen that when using strong inductive biases, such as the supervised losses in this experiment, the degree of disentanglement does not significantly affect the segmentation performance (utility).

\begin{figure}[b]
    \centering
    \vspace{-0.1in}
    \includegraphics[width=0.4\textwidth]{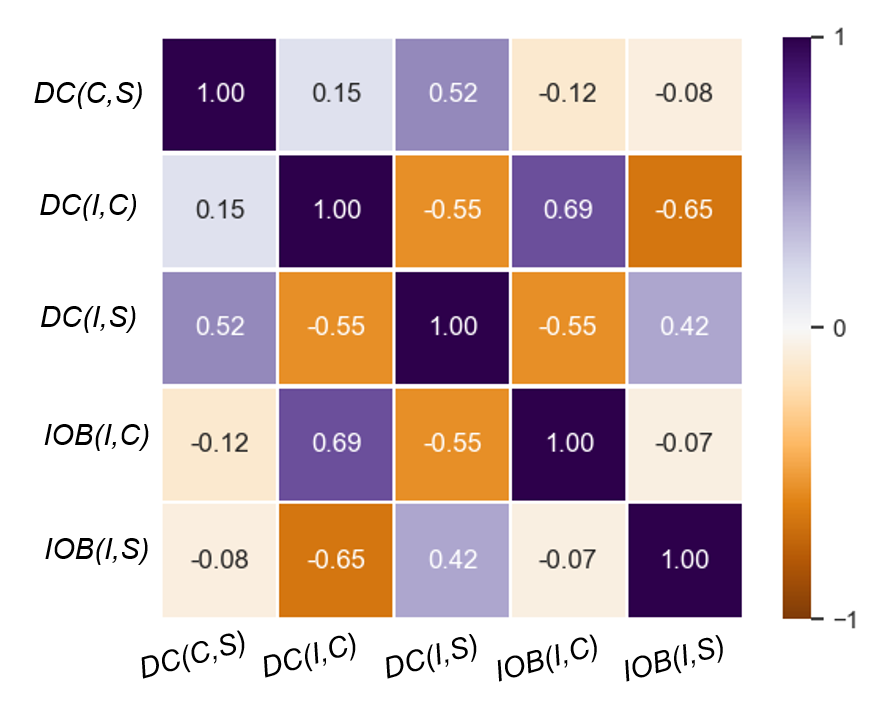}
    \vspace{-0.15in}
    \caption{Pearson correlation coefficients of the proposed metrics across all models visualized as a heatmap. Values close to 1 and -1 indicate a strong correlation.}
    \label{fig:total_corr}
\end{figure}

\begin{figure*}[t]
    \centering
    \includegraphics[width=\textwidth]{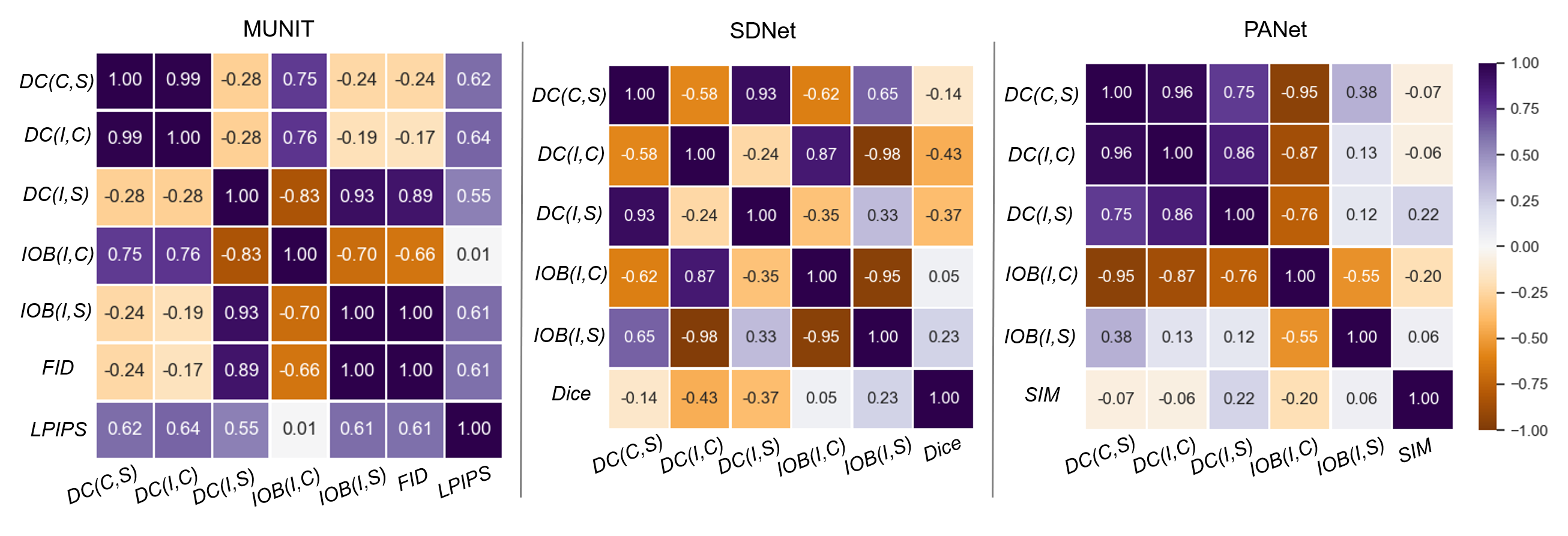}
    \caption{Pearson correlation of the proposed metrics across all applications/models visualized as heatmap. Values close to 1 and -1 indicate strong correlation.}
    \label{fig:per_task_corr}
\end{figure*}

\section{DeepFashion Description and PANet Training Setup}
\label{sec:panet_supp}
\textbf{Data.}
We use DeepFashion~\cite{liuLQWTcvpr16DeepFashion}, a large-scale dataset with over 800,000 diverse images of people in different poses and clothing, that also has annotations of body joints. We only used full-body images, specifically 32k images for training and 8k images for testing.

\textbf{Training setup.}
PANet is trained in an unsupervised way with the following loss function:
\begin{equation}
    \mathcal{L}_{total} = \lambda_1 \mathcal{L}_{rec} + \lambda_2 \mathcal{L}_{equiv},
\end{equation}
where $\mathcal{L}_{rec}$ is the mean absolute error between the reconstructed and the input image. $\mathcal{L}_{equiv}$ is an equivariance cost, that ensures that the mean and covariance of the parts coordinates don't change after some style transformation. Based on the implementation details presented in~\cite{lorenz2019unsupervised}, we set $\lambda_1=\lambda_2=1$. \\

\section{Metrics Correlation and Disentanglement-performance Trade-off}
\label{sec:corr}
As noted in Sec.~3, we report that the proposed metrics are uncorrelated with each other. Here, we present the Pearson correlation computed between disentanglement and performance metrics for each of the investigated models. Intuitively, contrary to the desired low (or no) correlation between disentanglement metrics across all models (see Fig.~\ref{fig:total_corr}), we would expect that the performance metric(s) of each application would be correlated with at least one $DC$ or $IOB$ variant. In fact, this correlation can be exploited to find the ``sweet spot" between disentanglement and performance. Fig.~\ref{fig:per_task_corr} confirms our intuition for all investigated models, highlighting the strong correlation of FID and LPIPS in the MUNIT scenario, which is the only model that utilizes both $C$ and $S$ directly in the main task, \textit{i.e.}~I2I translation, and not in any parallel one.

\section{Qualitative Evaluation}
\label{sec:qual}
We visualize the content and style representations in order to reason about their interpretability. We consider the content semantic if distinct objects appear in different channels, whereas the style is semantic when images reconstructed while traversing the style manifold between two points have smooth appearance changes, and are realistic. 

\begin{figure*}[t]
    \centering
    \includegraphics[width=0.9\textwidth]{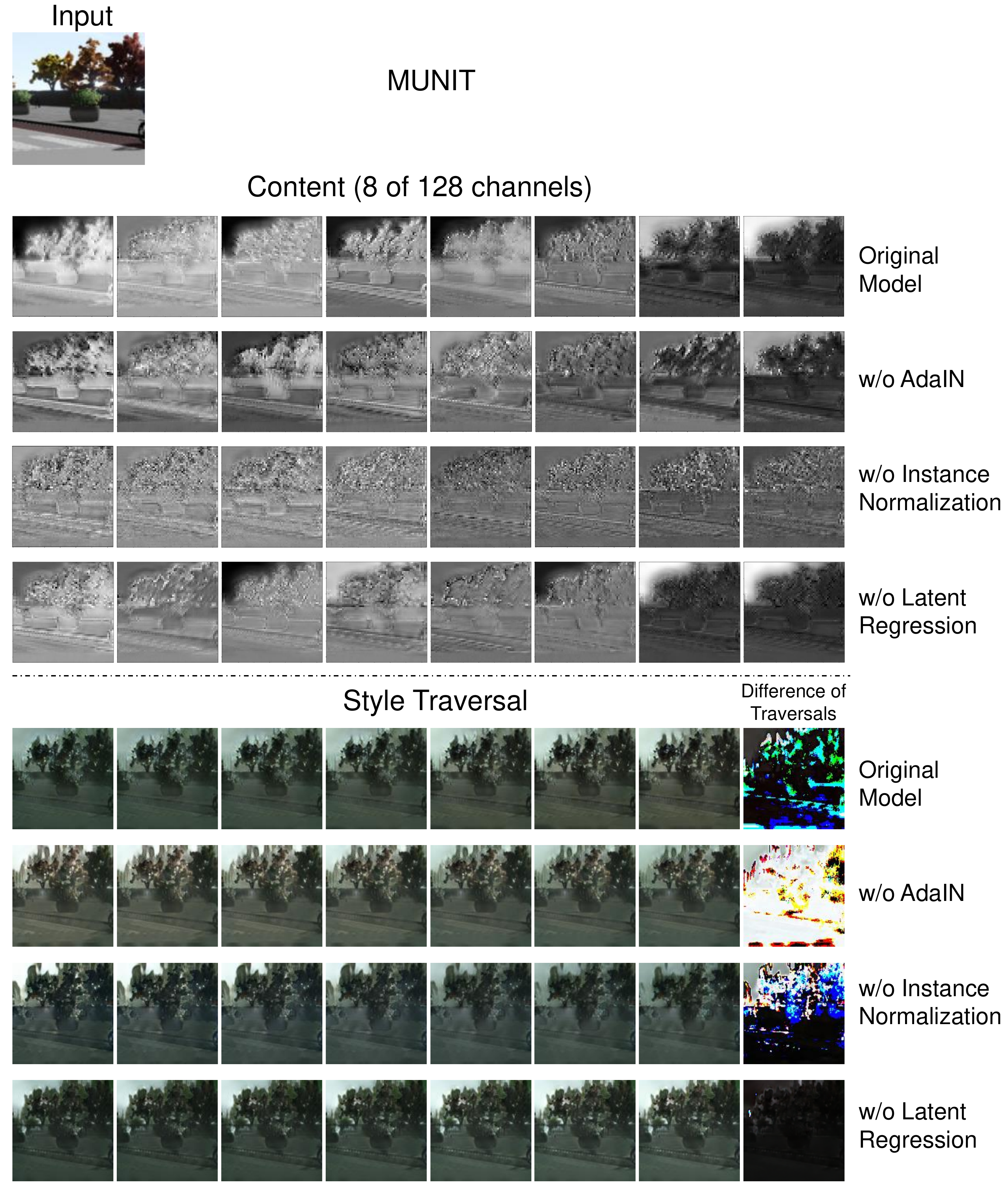}
    \caption{MUNIT: Qualitative examples to assess the interpretability of the content and style representations of the investigated model variants for different biases. For each variant, we show 8 channels of the content and 7 indicative style traversals and the difference between the first and last traversal images. The input image is depicted at the top left of the figure.}
    \label{fig:supp_munit}
\end{figure*}

\begin{figure*}[t]
    \centering
    \includegraphics[width=0.9\textwidth]{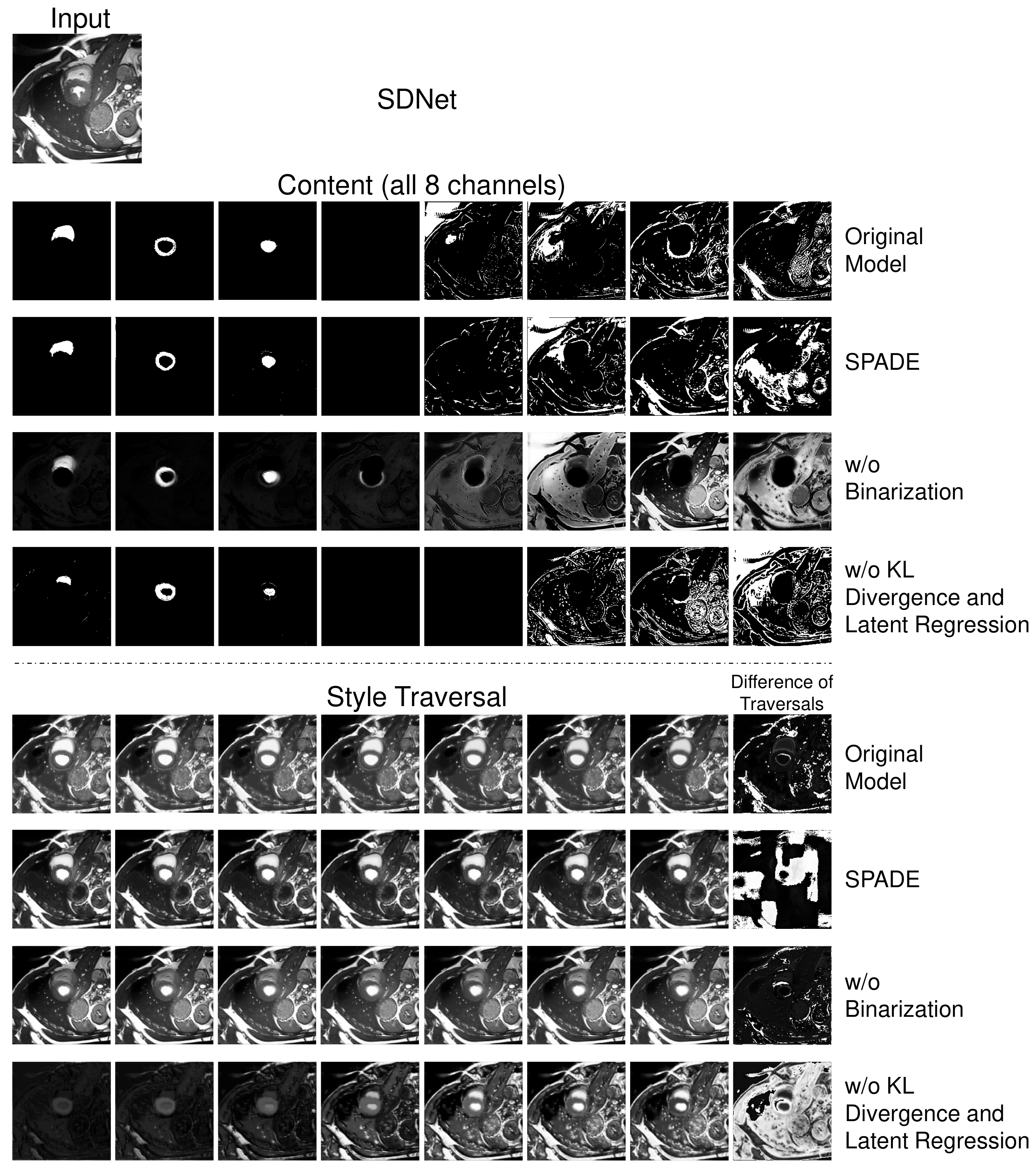}
    \caption{SDNet: Qualitative examples to assess the interpretability of the content and style representations of the investigated model variants for different biases. For each variant, we show 8 channels of the content and 7 indicative style traversals and the difference between the first and last traversal images. The input image is depicted at the top left of the figure.}
    \label{fig:supp_sdnet}
\end{figure*}

\begin{figure*}[t]
    \centering
    \includegraphics[width=0.9\textwidth]{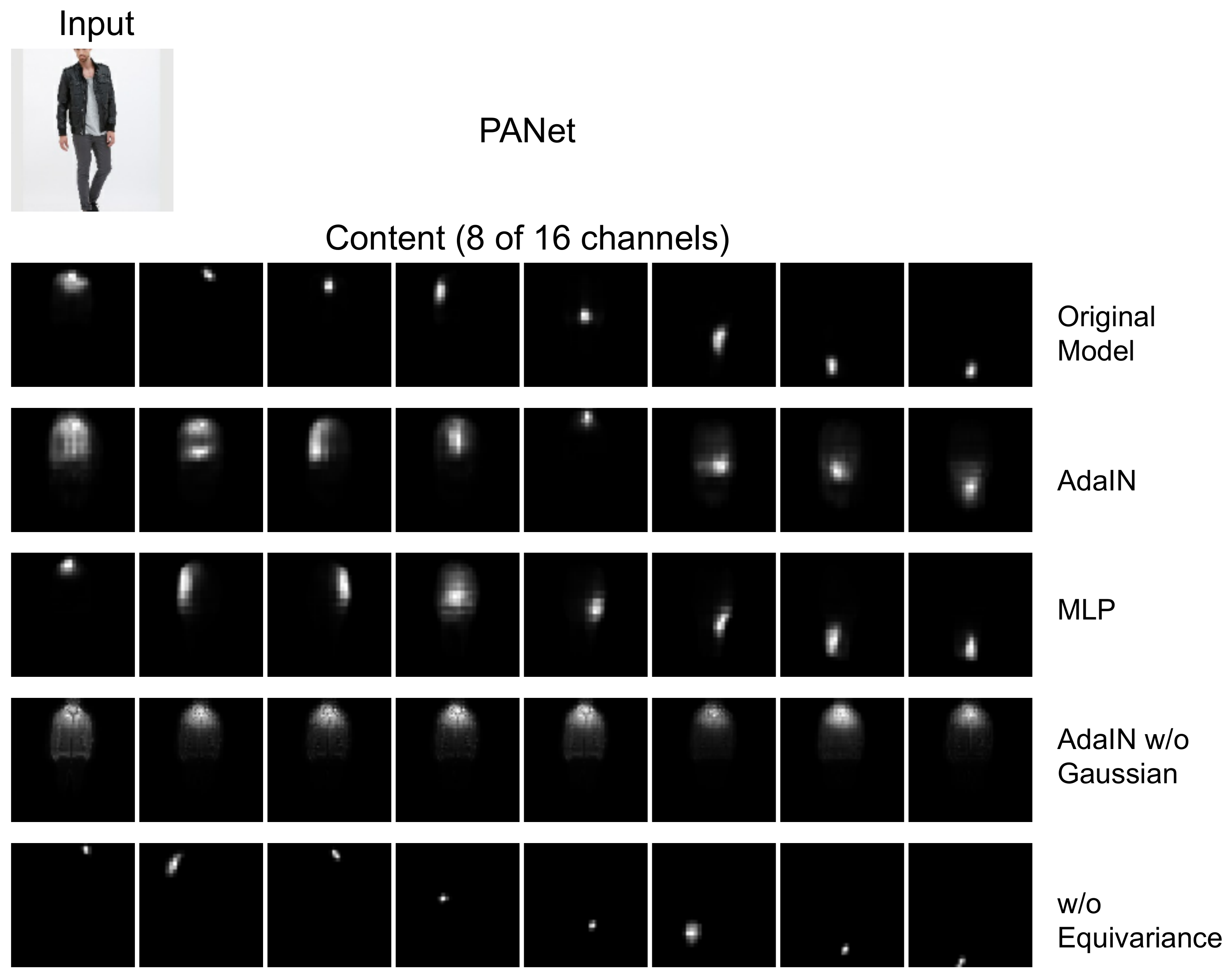}
    \caption{PANet: Qualitative examples to assess the interpretability of the content and style representations of the investigated model variants for different biases. For each variant, we show 8 channels of the content. Note that since PANet does not assume a prior distribution on the style, no style are shown. The input image is depicted at the top left of the figure.}
    \label{fig:supp_ppnet}
\end{figure*}

As an extension of the samples presented and discussed in Sec.~5.4, here we provide visualizations for all model variants. In particular, Figs.~\ref{fig:supp_munit} and \ref{fig:supp_sdnet} depict several channels of content, as well as style traversals for different MUNIT and SDNet model variants, respectively. However, Fig.~\ref{fig:supp_ppnet} presents solely content representations, as PANet does not assume a prior distribution on the style latent vector, thus style traversals are not possible. When interpolating between two style vectors, the originally proposed MUNIT produces realistic images, and smooth appearance changes. Instead, removing the LR constraint affects the image quality. Similarly, the original SDNet presents high image quality and smooth transitions, while removing the content Binarization leads to low intensity (style) diversity.

\end{document}